\documentclass{article}
     \PassOptionsToPackage{numbers, compress}{natbib}

     \usepackage[preprint]{neurips_2019}

\usepackage[utf8]{inputenc} 
\usepackage[T1]{fontenc}    
\usepackage{hyperref}       
\usepackage{url}            
\usepackage{booktabs}       
\usepackage{amsfonts}       
\usepackage{nicefrac}       
\usepackage{microtype}      
\usepackage{bm,amsmath,amsfonts,mathrsfs,amsthm}

\usepackage{graphicx}
\renewcommand{\eqref}[1]{Eq.~(\ref{#1})}

\usepackage{color}

\title{Deep Adversarial Belief Networks}

\author{%
  Yuming Huang$^1$, Ashkan Panahi$^2$, Hamid Krim$^2$, Yiyi Yu$^3$, Spencer L. Smith$^3$ \\
  $^1$Department of Physics, North Carolina State University\\
  $^2$Department of Electrical and Computer Engineering, North Carolina State University\\
  $^3$Department of Electrical and Computer Engineering, University of California Santa Barbara\\
  \texttt{yhuang26@ncsu.edu} \\
}

\begin{document}

\maketitle

\begin{abstract}
  We present a novel adversarial framework for training deep belief networks (DBNs), which includes replacing the generator network in the methodology of generative adversarial networks (GANs) with a DBN and developing a highly parallelizable numerical algorithm for training the resulting architecture in a stochastic manner. Unlike the existing techniques, this framework can be applied to the most general form of DBNs with no requirement for back propagation. As such, it lays a new foundation for developing DBNs on a par with GANs with various regularization units, such as pooling and normalization. Foregoing back-propagation, our framework also exhibits superior scalability as compared to other DBN and GAN learning techniques. We present a number of numerical experiments in computer vision as well as neurosciences to illustrate the main advantages of our approach.

\end{abstract}
\section{Introduction}
 An essential problem in statistical machine  learning (ML) is to model a given data set as a collection of independent samples from an underlying probability distribution. This distribution is generally referred to as a generative model. Representing and training generative models has a long and fruitful history and is popular in different mathematical modeling disciplines related to ML. The advent of deep learning and its associated methodology,  have in recent years resulted in remarkable advances in the area of inference. Generative Adversarial Networks (GANs) are among the chief examples, and have gained enormous attention in different domains of application, including automatic translation \cite{yu2017seqgan, isola2017image}, image generation \cite{goodfellow2014generative, reed2016generative,arjovsky2017wasserstein}and super-resolution \cite{ledig2017photo}. Deep Belief Networks (DBNs) are another set of highly popular examples with a wide range of application in acoustics \cite{mohamed2012acoustic,lee2009unsupervised}, computer vision \cite{nair20093d, teh2001rate} and others.

 Within the class  of generative models, the notion of generative network presents a marked  difference with  modern ML techniques,  such as GANs and DBNs, and other conventional methods. In a nutshell, a generative network consists of a randomized computational unit, which is able to generate random realizations from a wide variety of distributions. For instance, GANs utilize a standard deep neural network (DNN) that we refer to as  a neural generator, fed with a random sample from a fixed distribution. On the other hand, DBNs consist of a Markov chain of random vectors with the last vector in the chain as the output and the others as hidden features. In contrast to  conventional techniques, generative networks in GANs and DBNs encode the generative models in an implicit way, whereby the desired probability distribution may be computationally unfeasible, but can still   be statistically sampled efficiently. This fundamental difference leads to an incredible potential in representing highly complex generative models, with  however,  a radical paradigm shift in the  training methodology. In this respect, GANs provide a novel and numerically efficient training approach, relying on an adversarial learning framework and the stochastic gradient descent (SGD) technique in back-propagation.

It has been observed that in many cases, DBNs have remarkable advantages over the neural generators in GANs. A motivating example, considered in this work, is modeling the recorded activities of biological neurons from the visual cortex area of a mouse brain under visual stimulation. While neural generators inherit the limiting properties of  neural networks, such as continuity and differentiability, DBNs enjoy  much more versatile statistical properties, including sparsity and less severe regularity \cite{srivastava2012multimodal, boureau2008sparse}. For modeling biological neurons, sparsity considerably limits standard GANs' performance. Relying on a symmetric probabilistic relation between different layers, some DBNs can also be used in a reverse order, i.e. by feeding the data at the output layer and reversely generating the hidden features. DBNs based on Restricted Boltzmann Machines (RBMs) are prominent examples of such reversible networks. This unique property yields an efficient method for feature extraction, which has been exploited  in different applications such as data completion and denoising \cite{lee2009convolutional,lee2009unsupervised}. It has also been used in supervised learning problems such as classification, by learning a joint generative model for the data and the labels and feeding the data to generate the labels \cite{hinton2006fast,lee2009convolutional,chen2015spectral}. Using reversible DBNs, as we later argue in this paper,  also allows us to symmetrically capture the relation of  stimuli and neural activities, such that one of them can be inferred from the other. In addition, when given a small amount of training data, DBNs have the potential of  a better statistical performance than DNNs, on account of their Bayesian nature. This turns out to be  crucial to modeling neural activities as these recorded data are often in short supply.

Despite their numerous advantages, DBNs are less popular than GANs in practice, especially when highly deep structures are required to represent complex models. One reason is that unlike neural generators, the existing training techniques for DBNs are based on  layer-wise Gibbs sampling and/or variational methods, resulting in a substantially slower convergence rate than GANs. Moreover, the formulation of GANs admits various inference principles, as exemplified by the Wasserstein GAN (WGAN) architecture \cite{arjovsky2017wasserstein}, while DBNs are generally trained on the basis of the Maximum Likelihood (MaL) principle, which can become numerically unstable in many practical situations \cite{arjovsky2017wasserstein}. Furthermore, the various well known regularizers for neural generators such as pooling layers and normalization,  are not readily used in DBNs as of now.   Our goal in this paper is to address the afore-mentioned  issues by endowing  DBNs with similar training methodologies  available for GANs, while avoiding the conventional layer-wise training based on the MaL principle. The result may also be interpreted as a new generation of GANs with more flexible DBNs as their generators. We show that applying the adversarial learning approach to DBNs leads to a numerically more efficient algorithm than GANs, since unlike the back-propagation algorithm in GANs, our training approach is parallelizable over different layers. The main contributions of this paper can be summarized as follows:
\begin{itemize}
    \item Inspired by GANs, we develop an adversarial training framework for DBNs with superior numerical properties, including scalability through parallelization and compatibility with the acceleration and adaptive learning rate schemes.
    \item Unlike  existing approaches, our framework can address the most generic form of DBNs, creating a potential to incorporate similar regularization units as DNNs, such as pooling and normalization. Focusing on the standard RBM-based DBNs and convolutional DBNs, we leave the details of further generalizations to a future study. 
    
    \item Based on our framework, we develop algorithms that train DBNs under different metrics than the MaL principle, such as the Wasserstein distance.
    
    \item  We consider a number of illustrative experiments with the MNIST handwritten digits dataset \cite{lecun1998gradient} as well as the aforementioned biological neural activities.
\end{itemize}

\subsection{Related Literature}
DBNs belong to a broader family of Bayesian networks \cite{nielsen2009bayesian,neapolitan2004learning}. The most popular form of DBNs are for binary variables and generalize the "shallow" architecture of Restricted Boltzman Machines (RBMs) \cite{smolensky1986information, hinton2006reducing, ackley1985learning}. The most efficient methods of training RBMs, such the contrastive divergence method are based on  Monte Carlo (MC) sampling and variational Bayesian techniques \cite{tieleman2008training,sutskever2010convergence,hinton2012practical}. Similar techniques are used in a layer-wise fashion for training DBNs consisting of multiple layers of RBMs \cite{bengio2007greedy, boureau2008sparse, hinton2006fast}. Another approach for training DBNs is based on the variational lower bound (a.k.a Evidence Lower Bound ELBO) \cite{mnih2014neural}. Modifications of DBNs are also considered in the literature. In \cite{teh2001rate} and \cite{nair2010rectified} for example, the application of DBNs to non-binary variables is discussed by respectively introducing binomial and rectified linear (ReLU) units. To impose shift invariance, convolutional DBNs are introduced in \cite{lee2009convolutional}. Using DBNs for modeling neural activities has also been considered in \cite{lee2008sparse, zheng2014eeg}.
Compared to DBNs  GANs and neural generators belong to a more recent literature. The original idea of GANs stems from the original work in \cite{goodfellow2014generative}. Different variations of GANs, such as WGAN \cite{arjovsky2017wasserstein} and Deep Convolutional GANs (DC-GANs) \cite{radford2015unsupervised} are highly popular in the literature. Conditional GANs are introduced in \cite{mirza2014conditional} for modeling the relation of two variables such as images and labels \cite{gauthier2014conditional, isola2017image}. It is worth noting that using adversarial learning for training neural generators is not limited to GANs. For example, \cite{makhzani2015adversarial} introduces a training method based on the probabilistic auto encoder architecture used in the so-called Variational Auto Encoders (VAEs) \cite{kingma2013auto} and adversarial learning. 

\section{Mathematical Background}

\newcommand{\bGamma}{\bm{\Gamma}}
\newcommand{\bSigma}{\bm{\Sigma}}
\newcommand{\bOmega}{\bm{\Omega}}

\newcommand{\bdelta}{\bm{\delta}}
\newcommand{\bomega}{\bm{\omega}}
\newcommand{\bgamma}{\bm{\gamma}}
\newcommand{\bepsilon}{\bm{\epsilon}}
\newcommand{\blambda}{\bm{\lambda}}
\newcommand{\btheta}{\bm{\theta}}
\newcommand{\bpsi}{\bm{\psi}}
\newcommand{\bmeta}{\bm{\eta}}
\newcommand{\bzeta}{\bm{\zeta}}
\newcommand{\bmu}{\bm{\mu}}
\newcommand{\bnu}{\bm{\nu}}
\newcommand{\bpi}{\bm{\pi}}
\newcommand{\bsigma}{\bm{\sigma}}

\newcommand{\tilDelta}{\tilde{\Delta}}
\newcommand{\tlDelta}{\tilde{\Delta}}
\newcommand{\tlepsilon}{\tilde{\epsilon}}
\newcommand{\tltheta}{\tilde{\theta}}

\newcommand{\bA}{\mathbf{A}}
\newcommand{\bD}{\mathbf{D}}
\newcommand{\bE}{\mathbf{E}}
\newcommand{\bG}{\mathbf{G}}
\newcommand{\bH}{\mathbf{H}}
\newcommand{\bI}{\mathbf{I}}
\newcommand{\bJ}{\mathbf{J}}
\newcommand{\bL}{\mathbf{L}}
\newcommand{\bM}{\mathbf{M}}
\newcommand{\bN}{\mathbf{N}}
\newcommand{\bP}{\mathbf{P}}
\newcommand{\bQ}{\mathbf{Q}}
\newcommand{\bR}{\mathbf{R}}
\newcommand{\bS}{\mathbf{S}}
\newcommand{\bT}{\mathbf{T}}
\newcommand{\bW}{\mathbf{W}}
\newcommand{\bX}{\mathbf{X}}
\newcommand{\bY}{\mathbf{Y}}
\newcommand{\bZ}{\mathbf{Z}}

\newcommand{\ba}{\mathbf{a}}
\newcommand{\bb}{\mathbf{b}}
\newcommand{\bc}{\mathbf{c}}
\newcommand{\bd}{\mathbf{d}}
\newcommand{\be}{\mathbf{e}}
\newcommand{\mbf}{\mathbf{f}}
\newcommand{\bg}{\mathbf{g}}
\newcommand{\bh}{\mathbf{h}}
\newcommand{\bl}{\mathbf{l}}
\newcommand{\bn}{\mathbf{n}}
\newcommand{\bp}{\mathbf{p}}
\newcommand{\bq}{\mathbf{q}}
\newcommand{\br}{\mathbf{r}}
\newcommand{\bs}{\mathbf{s}}
\newcommand{\bu}{\mathbf{u}}
\newcommand{\bv}{\mathbf{v}}
\newcommand{\bw}{\mathbf{w}}
\newcommand{\bx}{\mathbf{x}}
\newcommand{\by}{\mathbf{y}}
\newcommand{\bz}{\mathbf{z}}

\newcommand{\hbeta}{\hat{\beta}}
\newcommand{\hmu}{\hat{\mu}}
\newcommand{\htheta}{\hat{\theta}}
\newcommand{\hsigma}{\hat{\sigma}}

\newcommand{\hg}{\hat{g}}
\newcommand{\hp}{\hat{p}}
\newcommand{\hr}{\hat{r}}
\newcommand{\hs}{\hat{s}}
\newcommand{\hx}{\hat{x}}

\newcommand{\hN}{\hat{N}}

\newcommand{\hbSigma}{\hat{\bm{\Sigma}}}

\newcommand{\hba}{\hat{\mathbf{a}}}
\newcommand{\hbs}{\hat{\mathbf{s}}}
\newcommand{\hbx}{\hat{\mathbf{x}}}
\newcommand{\hbv}{\hat{\mathbf{v}}}

\newcommand{\hbW}{\hat{\mathbf{W}}}

\newcommand{\dif}{\text{d}}

\newcommand{\bbC}{\mathbb{C}}
\newcommand{\bbE}{\mathbb{E}}
\newcommand{\bbR}{\mathbb{R}}
\newcommand{\bbN}{\mathbb{N}}
\newcommand{\bbZ}{\mathbb{Z}}

\newcommand{\calA}{\mathcal{A}}
\newcommand{\calB}{\mathcal{B}}
\newcommand{\calC}{\mathcal{C}}
\newcommand{\calD}{\mathcal{D}}
\newcommand{\calE}{\mathcal{E}}
\newcommand{\calF}{\mathcal{F}}
\newcommand{\calH}{\mathcal{H}}
\newcommand{\calL}{\mathcal{L}}
\newcommand{\calN}{\mathcal{N}}
\newcommand{\calM}{\mathcal{M}}
\newcommand{\calP}{\mathcal{P}}
\newcommand{\calS}{\mathcal{S}}
\newcommand{\calT}{\mathcal{T}}
\newcommand{\calV}{\mathcal{V}}
\newcommand{\calW}{\mathcal{W}}
\newcommand{\calX}{\mathcal{X}}
\newcommand{\calY}{\mathcal{Y}}

\newcommand{\calhL}{\mathcal{\hat{L}}}

\newcommand{\tlA}{\tilde{A}}
\newcommand{\tlC}{\tilde{C}}
\newcommand{\tlD}{\tilde{D}}

\newcommand{\tlv}{\tilde{v}}
\newcommand{\tls}{\tilde{s}}
\newcommand{\tlx}{\tilde{x}}

\newcommand{\barb}{\bar{b}}
\newcommand{\barm}{\bar{m}}
\newcommand{\barn}{\bar{n}}
\newcommand{\barr}{\bar{r}}
\newcommand{\bary}{\bar{y}}

\newcommand{\barC}{\bar{C}}
\newcommand{\barD}{\bar{D}}
\newcommand{\barH}{\bar{H}}
\newcommand{\barK}{\bar{K}}
\newcommand{\barL}{\bar{L}}
\newcommand{\barX}{\bar{X}}
\newcommand{\barW}{\bar{W}}

\newcommand{\barba}{\bar{\ba}}
\newcommand{\barbg}{\bar{\bg}}
\newcommand{\barbx}{\bar{\bx}}
\newcommand{\barby}{\bar{\by}}
\newcommand{\barbz}{\bar{\bz}}

\newcommand{\tlbA}{\tilde{\bA}}
\newcommand{\tlbE}{\tilde{\bE}}
\newcommand{\tlbW}{\tilde{\bW}}

\newcommand{\tlbv}{\tilde{\bv}}
\newcommand{\tlbx}{\tilde{\bx}}

\newcommand{\tc}{\text{c}}
\newcommand{\td}{{\text{d}}}

\newcommand{\bzero}{\mathbf{0}}
\newcommand{\bone}{\mathbf{1}}

\newcommand{\suml}{\sum\limits}
\newcommand{\minl}{\min\limits}
\newcommand{\maxl}{\max\limits}
\newcommand{\infl}{\inf\limits}
\newcommand{\supl}{\sup\limits}
\newcommand{\liml}{\lim\limits}
\newcommand{\intl}{\int\limits}
\newcommand{\bigcupl}{\bigcup\limits}
\newcommand{\bigcapl}{\bigcap\limits}

\newcommand{\opconv}{\text{conv}}

\newcommand{\eref}[1]{(\ref{#1})}

\newcommand{\sinc}{\text{sinc}}
\newcommand{\tr}{\text{Tr}}
\newcommand{\var}{\text{Var}}
\newcommand{\cov}{\text{Cov}}
\newcommand{\tth}{\text{th}}
\newcommand{\proj}{\text{proj}}

\newcommand{\nwl}{\nonumber\\}

\newenvironment{vect}{\left[\begin{array}{c}}{\end{array}\right]}
\newtheorem{theorem}{Theorem}
\newtheorem{lemma}{Lemma}

\subsection{Problem Formulation: Training Deep Belief Networks}
Given an observed data set $\barX=\{x_n\}_{n=1}^N$ with $N$ data points $x_n$ from a data space (domain) $\calX$, we are to estimate a probability measure \footnote{To define such a measure, we naturally assume that $\calX$ is also equipped with a proper sigma algebra. Indeed, we are practically concerned only with the case where $\calX=\bbR^m$ is the space of $m-$dimensional real vectors with the standard Borel sigma algebra.}$\mu$ on $\calX$, with $\barX$ as its set of independent random samples. DBNs address this problem by generating a random variable with a desired distribution $\mu$. To this end,  multiple layers  $h^1,h^2,\ldots,h^L$ of random variables are considered. In the RBM-based DBNs, the $l^\tth$ layer $h^l=(h_1^l,h^l_2,\ldots,h^l_{d_L})$ is a random $d_L-$dimensional binary vector and the joint probability density function\footnote{For simplicity, we interchangeably use the terms probability distribution and probability mass.} (p.d.f) of the layers is written as
\begin{equation}\label{eq:Markov}
    \log p(h^1,h^2,\ldots,h^L)=\log C+ \suml_{l=1}^{L-1}\suml_{(i,j)\in [d_l]\times[d_{l+1}]}h^{l+1}_iw^l_{ij}h^l_j+\suml_{l=1}^{L}\suml_{i\in [d_l]}b_ih^l_i,
\end{equation}
where $w_{ij}^l,b_i^l$ are a set of weights and $C$ is a proper normalization constant. The output layer $h^L$ thus reflects the desired distribution, while the layers $\{h^l\}$ form a Markov chain. This is apparent in the RBM-based formulation in \eqref{eq:Markov} as the pdf can be factored by terms including only adjacent layers. We may write the joint distribution in the "forward" form $p(h^1,h^2,\ldots,h^L)=p_1(h^1)p_2(h^2\mid h^1)\ldots p_L(h^L\mid h^{L-1})$, where
\[
\log p_1(h^1)=\log C_1+\suml_{i\in d_1}b_i^1h_i^1,
\]
and with a proper choice of the constant $C_1$, is the marginal distribution of the input layer $h^1$ and
\[
\log p_l(h^l\mid h^{l-1})=\log C_l+{\suml_{(i,j)\in [d_{l-1}]\times[d_{l}]}h^{l}_iw^{l-1}_{ij}h^{l-1}_j+\suml_{i\in d_l}b_i^lh_i^l}
\]
are the transitional probabilities between the $l,l-1$ layers, with a suitable normalization constant $C_l$. This forward representation enable us to conveniently sample the output of DBNs by first sampling the input (by $p_1$) and by successively sampling the next layers (by $p_l$), given the realizations of the previous layers, until reaching the output. We observe that in the forward representation, the elements ${h_i^1}$ of the input layer are independent. Conditioned on their previous layer, the elements of the next layers are also independent. As the variables in \eqref{eq:Markov} are binary, the constants $C_l$ can also be explicitly calculated, resulting in  logistic functions for the probability of individual elements in the first layer as well as the conditional probabilities of the subsequent layers:
\begin{equation}\label{eq:logistic}
p_1\left(h_i^1=1\right)=\sigma\left(b_i\right),\quad p_l\left(h_i^l=1\mid h^{l-1}\right)=\sigma\left(b_i^l+\suml_{j\in [d_{l-1}]}w_{ij}h_j^{l-1}\right),
\end{equation}
where $\sigma(x)=1/(1+e^{-x})$ is the logistic function. Once a DBN is trained, it can also be used in a backward way by factorizing the joint pdf as $p_L(h^L)p_{L-1}(h^{L-1}\mid h^L)\ldots p_1(h^1\mid h^2)$, where with an abuse of notation we also denote the backward transitional probabilities by $p_l$. For the RBM-based model in \eqref{eq:Markov}, this factorization can be easily carried out, leading to similar expressions of \eqref{eq:logistic}, which allows us to reversely sample a set of hidden features from a give data point at the output layer.

DBNs are trained based on the Maximum Likelihood (MaL) principle. For a given set $\theta=\{\{w_{ij}^l\},\{b_i^l\}\}$ of weights, denote the marginal distribution of the output $h^L$ by $p_\theta(h^L)$. Then the MaL principle leads to the following optimization problem for training DBNs:
\begin{equation}\label{eq:ML}
    \minl_{\theta}-\suml_{n=1}^N\log p_\theta(h^L=x_n).
\end{equation}
The major difficulty  in \eqref{eq:ML} is the calculation of  the term $p_\theta(\ldotp)$, and its derivative is extremely difficult to calculate and making gradient-based optimization techniques not directly applicable. 
To overcome this difficulty, and make DBNs training more viable, we exploit in this paper,   the GANs methodology to lift the numerical difficulties  of the MaL-based optimization framework. 
\subsection{Proposed Method: Deep Adversarial Belief Networks}
For training the DBNs, we adopt a similar solution to GANs. We 
consider the empirical measure $\hmu=\frac 1 N\sum_n\delta_{x_n}$ of the data set $\barX$, where $\delta_x$ denotes Dirac's delta measure at point $x\in\calX$ and take the solution of the following optimization problem: 
\begin{equation}\label{eq:frame}
    \minl_{\mu\in\calM}D(\hmu,\mu),
\end{equation}
where $D$ is a positive distance or divergence function between two measures, and $\calM$ is the set of probability measures $p_\theta(h^L)$ on $\calX$ generated by the DBNs in \eqref{eq:Markov}
. We observe that \eqref{eq:frame} generalizes the MaL framework in \eqref{eq:ML}, since the latter is obtained as a special case, by letting $D$ be the Kullback Leibler divergence, i.e. $D(\hmu,\mu)=KL(\hmu||\mu)$. More generally and similarly to GANs, we consider those distance (or divergence) functions $D$ that can be written as
\begin{equation}\label{eq:divergence}
    D(\hmu,\mu)=\maxl_{f\in\calF}\bbE_{X\sim\hmu}[\phi(f(X))]+\bbE_{Y\sim\mu}[\psi(f(Y))],
\end{equation}
where $\calF$ is a family of real-valued functions on $\calX$, known as the \emph{discriminators}. Furthermore, $\phi,\psi$ are two real functions and the notation $\bbE_{X\sim\hmu}[\ldotp],\bbE_{Y\sim\mu}[\ldotp]$ implies that the variables $X,Y$ in the arguments of expectation are respectively distributed according to  $\hmu,\mu$. The original GAN formulation  uses $\phi(y)=\log(y)$ and $\psi(y)=\log (1-y)$ with $\calF$ as the set of all measurable functions, corresponding to the Jensen-Shannon divergence. The WGAN formalism is obtained by taking $\phi(x)=x$, $\psi(x)=-x$ and $\calF$ as the set of all 1-Lipschitz functions, which leads to the Wasserstein distance between measures. The MaL framework in \eqref{eq:ML} can also be obtained by setting $\phi(x)=\log(x)$ and $\psi(x)=-x$. In practice, the discriminator $f\in\calF$ is limited to the family of deep neural networks (DNNs) with a suitable fixed architecture. In this case, we denote the discriminator $f$ by $f_\rho$ where $\rho$ denotes the set of weights in the neural network at hand. 
Plugging \eqref{eq:divergence} into \eqref{eq:frame} and using the above-mentioned specifications of the discriminator, we obtain the following optimization framework:
\begin{equation}\label{eq:GAN}
    \minl_{\theta}\maxl_\rho \bbE_{X\sim\hmu}[\phi(f_\rho(X))]+\bbE_{Y\sim\mu=p_\theta}[\psi(f_\rho(Y))].
\end{equation}
Our proposed technique for training DBNs is hence entails solving the optimization problem in \eqref{eq:GAN} to obtain the set $\theta$ of parameters of the underlying DBN. As we shortly elaborate, the stochastic gradient method provides a practical scheme for this purpose.
We also observe that \eqref{eq:GAN} bears a similar adversarial interpretation to GANs:
As the loss function reflects the objective of the discriminator in distinguishing the "real" samples $X$ from the "fake" ones $Y$, the goal of the DBN ($p_\theta$) is to deceive the discriminator by counterfeiting  "true samples" in the best possible way.
\subsubsection{Algorithmic Details}
The optimization problem in \eqref{eq:GAN} can be solved by the SGD method: At each iteration $t=1,2,\ldots$ a set of samples (mini-batch) from either the data set $\barX$ or the output $Y=h^L$ of the underlying DBN is randomly selected. The gradient $g^t$ with respect to both $\theta$ and $\rho$ of their corresponding term $\bbE[\phi(f_\rho(X))]$ or $\bbE[\psi(f_\rho(Y))]$ in the objective of \eqref{eq:GAN},  are estimated using the samples, and subsequently applied. When considering the $t^\tth$ iteration, if a set of $b$ samples $x_1^t,x_2^t,\ldots, x^t_b$ from the data set $\barX$ is used, we adopt the standard procedure of estimating the gradient by calculating the sample mean:
\begin{equation}\label{eq:grad1}
\hg_t=\left[0,\ \frac 1b\suml_{i=1}^b\frac{\partial\phi\left(f_\rho\left(x_i^t\right)\right)}{\partial\rho}\right],
\end{equation}
where the estimate $\hg_t$ respectively includes the gradients with respect to $\theta$ and $\rho$ in the first and second entries. Note that the gradient with respect to $\theta$ is zero in this case. The above solution is not applicable when the DBN samples $y_1^t,y_2^t,\ldots,y_b^t$ are employed, since the relation of their corresponding term $\bbE[\psi(f_\rho(Y))]$ to $\theta$ is implicit in the underlying distribution $p_\theta$. For this reason, we first express the exact gradient of this term with respect to $\theta$ as\footnote{In the continuous variable case, the summation will be replaced by an integral, but the final expression remains unchanged.} 
\begin{eqnarray}\label{eq:trick}
&\frac{\partial\bbE[\psi(f_\rho(Y))]}{\partial\theta}=\frac{\partial}{\partial \theta}\suml_{\bh}\psi(f_\rho(h^L))p_\theta(\bh)
\nwl
&=\suml_{\bh}\psi(f_\rho(h^L))p_\theta(\bh)\frac{\partial}{\partial\theta}\log p_\theta(\bh)=\bbE_{\bh\sim p_\theta}\left[\psi(f_\rho(h^L))\frac{\partial}{\partial\theta}\log p_\theta(\bh)\right],
\end{eqnarray}
where $\bh=(h^1,\ldots,h^L)$, and  the notation $p_\theta(\bh)$ is used to refer to the joint p.d.f in \eqref{eq:Markov}. We observe in \eqref{eq:trick} that the expected value on the right hand side is over all layers in $\bh$, while the original expression on the left hand side is over the output $Y=h^L$. Next, we estimate the right hand side by generating $b$ samples $\bh^t_1,\bh^t_2,\ldots,\bh^t_b$ of the entire network, where $\bh^t_i=(h^{1,t}_i,h^{2,t}_i,\ldots,h^{L,t}_i)$ with $y_i^t=h_i^{L,t}$ as the $i^\tth$ sample of the output layer, and calculating the sample mean. This leads to the following expression for the gradient
\begin{equation}\label{eq:grad2}
\hg_t=\left[\frac 1b\suml_{i=1}^b\psi\left(f_\rho\left(y_i^t\right)\right)\frac{\partial\log p_\theta(\bh_i^t)}{\partial\theta},\ \frac 1b\suml_{i=1}^b\frac{\partial\psi\left(f_\rho\left(y_i^t\right)\right)}{\partial\rho}\right].
\end{equation}
We notice that the term $\frac{\partial\log p_\theta(\bh_i^t)}{\partial\theta}$ can be efficiently calculated on account of the Markovian properties of the DBNs, which allows us to efficiently express $\log p_\theta(\bh_i^t)=\log p_1(h^1)+\sum_l\log p_l(h^{l}\mid h^{l-1})$. This shows that the gradient of the variables at individual layers can be independently calculated in parallel, thus foregoing the back-propagation algorithm. This represents a great numerical advantage of adversarial DBNs over DNNs. We observe that for the RBM-based DBNs, calculating the term $\frac{\partial\log p_\theta(\bh_i^t)}{\partial\theta}$ amounts to differentiating the expressions in \eqref{eq:logistic}, which can be found in the standard literature of DBNs \cite{hinton2006fast}, and is hence skipped herein for space sake. Once the elements of the stochastic gradient $\hg$ are calculated based on either \eqref{eq:grad1} or \eqref{eq:grad2}, they are applied to their corresponding parameters with a suitable learning rate.
\subsubsection{Extensions}
Our training method by \eqref{eq:grad1}  and \eqref{eq:grad2} enables us to extend the existing framework of DBNs in multiple respects:
\\
{\bf Modifying MaL Principle}: We can easily alter our training principle by modifying the pair of functions $\phi,\psi$. In particular, we consider the Wasserstein metric and the JS divergence in our next experiments, which are popular choices in the GAN literature.
\\
{\bf Non-RBM Layers}: As seen, our training technique  is applicable to any DBN, such as \eqref{eq:logistic}, for which the derivative of the forward representation is simple to compute. For example, we may simply obtain the convolutional DBNs by replacing the linear terms $\sum w_{ij}^lh_j^{l-1}$ in \eqref{eq:logistic} with a convolution. The resulting expressions and derivatives are similar to those in \cite{lee2009convolutional} and are hence skipped, we nevertheless use the resulting algorithm in our experiments. Further operations such as normalization factors and pooling can also be incorporated in the description of the transitional probabilities $p_l(h^l\mid h^{l-1})$ in \eqref{eq:logistic}, and their adoption is postponed to a future work as they will impact the reversibility property. 
\\
{\bf Accelerated Learning:} Another advantage of our training methodology is that it admits standard techniques in optimization algorithms, such as acceleration and adaptive step size to improve convergence. We examine some of these approaches in our experiments. 
\section{Experiments}
In this section, we examine our proposed training algorithm by a way of two groups of numerical experiments. The first group concerns the application of DBNs to a computer vision problem, namely the MNIST dataset, containing 60,000 labeled samples of gray-scale hand-written digits for training and 10,000 more for testing. The second group investigates DBNs for modeling neural activities of the visual cortex under given visual stimuli. 
\subsection{Generation of MNIST-Like  Digits}
\label{sec:mnist1}
 The goal of our first experiment is to generate synthetic handwritten digits by a DBN, with or without control over the generated digit. For this experiment, we use the dataset of \cite{salakhutdinov2}, containing 1797 samples of  8 by 8 cropped MNIST images, further binarized by thresholding the original grey scale images. In the first part of this experiment, a DBN is adversarially trained by \eqref{eq:grad1} and \eqref{eq:grad2} in an unsupervised way, i.e. by feeding the image samples as the last layer $h^L$ and discarding the labels. Sampling the resulting DBN generates handwritten instances with no control over the underlying digit. In the second part, another DBN is trained in a supervised way (with ground truth labels) to gain control over the generated digit. For this purpose, we adopt a similar approach to the conditional GAN structure \cite{mirza2014conditional} by treating a pair of label and image as a data point $x_k$, which are respectively fed to the first ($h^1$) and last ($h^L$) layer of a DBN. The discriminator $f_\rho$ assumes this pair as an input and \eqref{eq:grad1} and \eqref{eq:grad2} are similarly used. The output of the discriminator in the two parts of our experiment can be interpreted as the likelihood of the input samples following the distribution of the training images, either unconditionally or given a label. 
 
 \begin{figure}
    \centering
    \includegraphics[width=13.5cm]{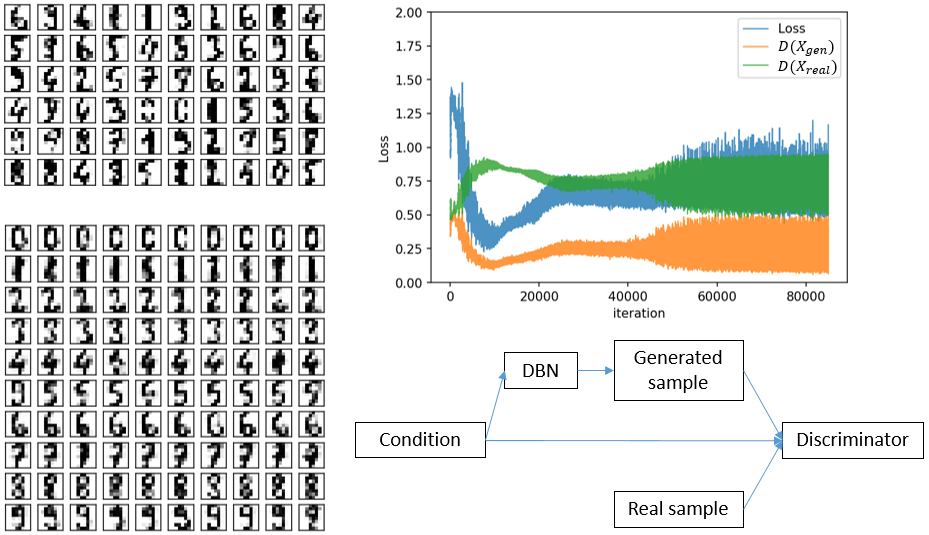}
    \caption{Top-left: random generated samples from DBN trained using the unsupervised framework of GAN. Top-right: loss function and discriminator score of real and generated samples over a number of training iterations. Bottom-left: random generated samples from the conditional DBN, where the conditional input for each row represents digit from 0 to 9. Bottom-right: a diagram of the training framework for the conditional DBN.}
    \label{fig:mnist}
\end{figure}

 In both parts of our experiment, the discriminator is a 2-layer densely connected neural network, whose input vector length, the hidden layer, and the output layer respectively are [64,64,1]. The output layer has a sigmoid activation function. In the first part, we employ a
densely connected 3-layer DBN as in \eqref{eq:logistic}, and each with  64 units.  For the second part, we add 
an input layer of length 10, corresponding to the one-hot encoding (converting categorical integers to a binary vector) of the ten digit labels.
 
 The diagram of the DBN and the discriminator are shown in Fig. \ref{fig:mnist}. Samples of the generated images in the two experiments are also shown in Fig. \ref{fig:mnist}. Note that the last layer of the resulting DBNs generates binary images of the digits, while the values of the conditional distribution $p_L(h_i^L\mid h^{L-1})$ of the pixels of the last layer given the previous layer in \eqref{eq:logistic}, can be used as the grey scale images of the original MNIST digits, before thresholding. We show the results of the conditional distribution of the last layer in Fig. \ref{fig:mnist}. 
  We observe that the generated digits are well distributed and resemble the real digits. Moreover, in the bottom plot, each row is conditioned on a certain label, which shows that the generation of the digits is to a large extent, associated with the labels, while the digits maintain variability for the same label.





\begin{figure}
    \centering
    \includegraphics[width=13.5cm]{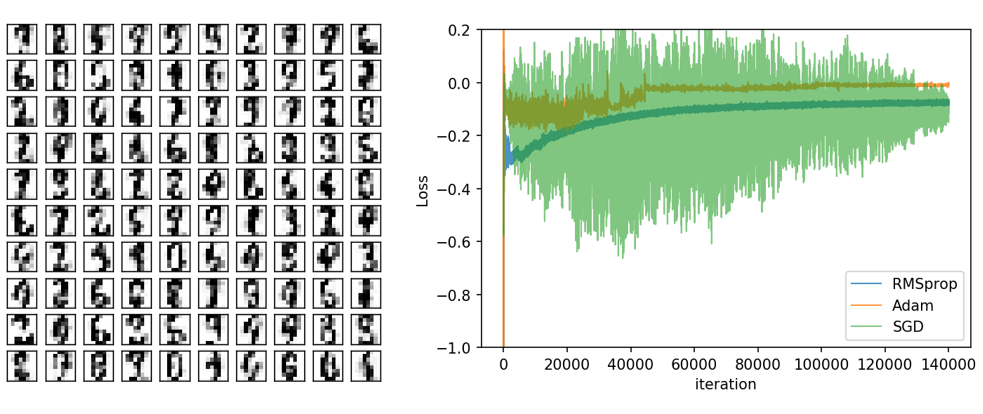}
    \caption{Left: random samples from DBN trained using the framework of WGAN. Right: loss function over a number of training iterations, with the discriminator trained by different optimizers at the same learning rate.}
    \label{fig:wgan_mnist}
\end{figure}

{\it Training by Wasserstein distance:} We also conducted the first part of our experiment using the WGAN formalism. Specifically, we apply weight clipping \cite{arjovsky2017wasserstein} on the discriminator, remove the sigmoid function at the end of the discriminator, and change the loss function elements to $\phi(x)=x,\psi(x)=-x$ in \eqref{eq:grad1} and \eqref{eq:grad2} according to the Wasserstein distance. We show in Fig. \ref{fig:wgan_mnist} some generated samples of the digits using Wasserstein DBN, which demonstrates that training DBNs with different metrics than the KL divergence leads to different distributional properties of the generated images. In this experiment, we also explore different adaptive learning rate strategies. The convergence properties of three different optimizers, namely SGD, RMSprop \cite{tieleman2012lecture} and Adam \cite{kingma2014adam} are depicted on the right hand side of Fig. \ref{fig:wgan_mnist}. As seen, SGD leads to highest variance and slowest convergence, while Adam results in the most suitable saddle point solution of \eqref{eq:GAN}. 

\subsection{Classification of MNIST}
In this experiment, we use the DBN formalism for classification of MNIST images. For this purpose, we train a conditional convolutional DBN on the original MNIST dataset using the proposed framework. This is similar to the second part of our previous experiment with digits generation, with a slight difference that the DBN takes images as a non-binary input and outputs a vector of length 10. We repeat this experiment by binarizing the entire MNIST dataset, but since the two results are similar, we only present one set of results. The images are still used as the conditional inputs to the discriminator, as in Fig. \ref{fig:mnist}. The DBN includes 4 convolutional layers, with the following number of filters [32,32,16,10], and their corresponding  sizes [11,11,5,4]. In the discriminator architecture, we first feed the conditional image to a CNN to generate a feature vector of length 64. We subsequently concatenate the feature vector with the DBN's output (generated labels) and pass the result through a two layer linear neural network.

Since generative models are not specialized for  classification, their training is usually followed by a fine tuning stage, where their weights are updated by backpropagation as a conventional CNN \cite{lee2009unsupervised}. We use the first layer of a trained convolutional DBN as the pretrained weights for the first layer filters of a CNN, and fine-tune the weights using back-propagation. The classification performance is compared with direct training of a CNN classifier of the same structure. The resulting accuracy for different sizes of the training set are listed in Table \ref{tab:mnist}. This shows that CNN initialized by DBN outperforms normal CNN for a small training set and has better generalizability. For the CNN without DBN pre-training and a small size of the training set, the result is highly unstable. In contrast, CNN initialized by DBN exhibits a considerably more consistent performance. The reported accuracy for CNN is the average of 100 runs. 

\begin{table}[]
    \centering
    \begin{tabular}{|c|c|c|c|}
    \hline
    training set size & DBN + CNN & CNN \\ \hline
    100              &      77.772\% (1.501\%)         &  70.037\% (6.32\%)   \\ \hline
    1000             &      96.081\% (0.622\%)         &  94.625\% (0.355\%)   \\ \hline
    60000            &   99.140\% (0.096\%)            & 98.906\% (0.092\%)    \\ \hline
    \end{tabular}
    \caption{Classification accuracies when number of training samples is 100, 1000 and 60000 respectively, averaged over 100 trials, with standard deviation in the parentheses. Compared with CNN, the use of DBN as pretraining shows better generalizability for small training set. }
    \label{tab:mnist}
\end{table}


\subsection{Modeling Visual Cortex Neural Activities}
Modelling neural spikes is a natural application of DBNs \cite{lee2008sparse}.  
The power of DBNs in addressing the limitations of DNNs,  such as sparsity and reduced datasets, is demonstrated in this task. They naturally generate binary outputs, which can be easily interpreted as neural spikes. In this experiment, we show that a DBN is capable of modeling sparse spike signals.

 Our dataset is recorded by a two-photon calcium imaging system capturing large scale neural activities \cite{stirman2016wide,huang2018}. We simultaneously record activities of individual neurons as time series in the primary visual cortex (V1) and the anterolateral (AL) areas of the visual cortex of an awake mouse. Top 50 neurons whose activities are most correlated across 20 trials are selected for modeling, and their binary spike trains are obtained by applying a standard deconvolution technique to the recorded time series.

\begin{figure}
    \centering
    \includegraphics[width=13.5cm]{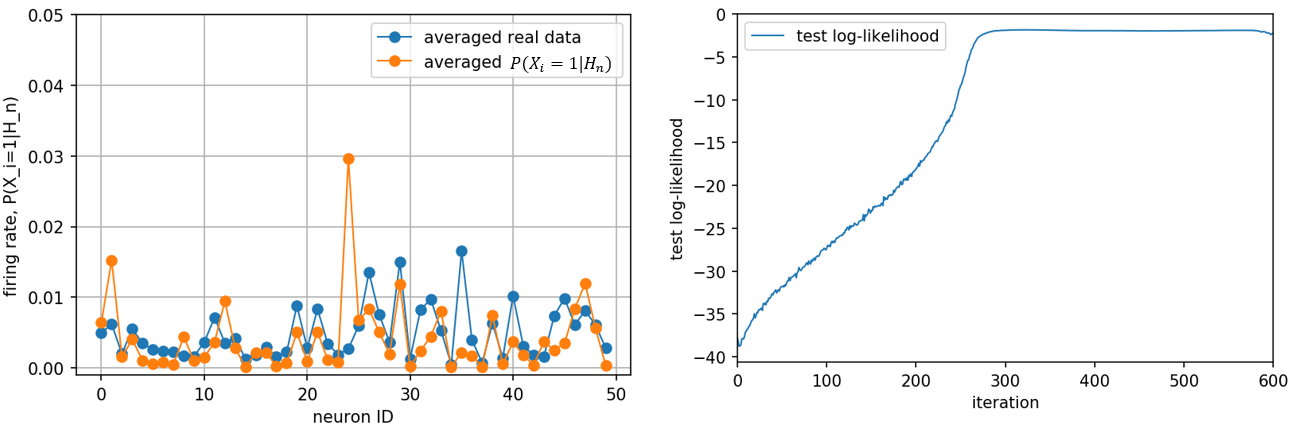}
    \caption{Left: real firing rate of each neuron and the firing rates given by the DBN. Right: log-likelihood of the real data on the DBN during training.}
    \label{fig:spikes}
\end{figure}

We model the spikes of the 50 neurons independent of the visual stimuli with a four-layer dense DBN. The number of units per layer are [128,128,128,50]. The discriminator is a two-layer dense neural network with 64 neurons in the hidden layer. The training procedure is similar to the first part of the experiment in Section \ref{sec:mnist1}. In Fig. \ref{fig:spikes}, the firing probabilities for individual neurons, given by the DBN are depicted, which exhibit high resemblance to the real firing rates, despite the fact that the overall firing rate is very low (about 1 spike per 100 frames).  We also verify that the log-likelihood of the real data on DBN increases during training. The likelihood of the output layer of the DBN is estimated by sampling $p_L(h^L\mid h^{L-1})$, the distribution of the last layer conditioned on the penultimate layer, and averaging, amounting to the total probability rule. 

\section{Conclusion}
In this paper, we proposed an adversarial training framework for DBNs. The experiments verify that our method works under different structures and settings, including GAN, WGAN, conditional GAN and various optimizers. The development of this method opens a promising way to train complex DBNs. Future works include implementing more components for DBN, improving the modeling capability and training stability, and explore more complex and dedicated DBN structures for neural modelling, such as 3D convolution and recurrent structures. 

\bibliographystyle{plain}
\bibliography{bibfile}

\begin{thebibliography}{10}

\bibitem{ackley1985learning}
David~H Ackley, Geoffrey~E Hinton, and Terrence~J Sejnowski.
\newblock A learning algorithm for boltzmann machines.
\newblock {\em Cognitive science}, 9(1):147--169, 1985.

\bibitem{arjovsky2017wasserstein}
Martin Arjovsky, Soumith Chintala, and L{\'e}on Bottou.
\newblock Wasserstein gan.
\newblock {\em arXiv preprint arXiv:1701.07875}, 2017.

\bibitem{bengio2007greedy}
Yoshua Bengio, Pascal Lamblin, Dan Popovici, and Hugo Larochelle.
\newblock Greedy layer-wise training of deep networks.
\newblock In {\em Advances in neural information processing systems}, pages
  153--160, 2007.

\bibitem{boureau2008sparse}
Y-lan Boureau, Yann~L Cun, et~al.
\newblock Sparse feature learning for deep belief networks.
\newblock In {\em Advances in neural information processing systems}, pages
  1185--1192, 2008.

\bibitem{chen2015spectral}
Yushi Chen, Xing Zhao, and Xiuping Jia.
\newblock Spectral--spatial classification of hyperspectral data based on deep
  belief network.
\newblock {\em IEEE Journal of Selected Topics in Applied Earth Observations
  and Remote Sensing}, 8(6):2381--2392, 2015.

\bibitem{gauthier2014conditional}
Jon Gauthier.
\newblock Conditional generative adversarial nets for convolutional face
  generation.
\newblock {\em Class Project for Stanford CS231N: Convolutional Neural Networks
  for Visual Recognition, Winter semester}, 2014(5):2, 2014.

\bibitem{goodfellow2014generative}
Ian Goodfellow, Jean Pouget-Abadie, Mehdi Mirza, Bing Xu, David Warde-Farley,
  Sherjil Ozair, Aaron Courville, and Yoshua Bengio.
\newblock Generative adversarial nets.
\newblock In {\em Advances in neural information processing systems}, pages
  2672--2680, 2014.

\bibitem{hinton2012practical}
Geoffrey~E Hinton.
\newblock A practical guide to training restricted boltzmann machines.
\newblock In {\em Neural networks: Tricks of the trade}, pages 599--619.
  Springer, 2012.

\bibitem{hinton2006fast}
Geoffrey~E Hinton, Simon Osindero, and Yee-Whye Teh.
\newblock A fast learning algorithm for deep belief nets.
\newblock {\em Neural computation}, 18(7):1527--1554, 2006.

\bibitem{hinton2006reducing}
Geoffrey~E Hinton and Ruslan~R Salakhutdinov.
\newblock Reducing the dimensionality of data with neural networks.
\newblock {\em science}, 313(5786):504--507, 2006.

\bibitem{huang2018}
Yuming Huang, Ashkan Panahi, Han Wang, Bo~Jiang, Hamid Krim, Yiyi Yu, Spencer
  L.~Smith, and Liyi Dai.
\newblock Model-free inference of neuronal connectivity via embedding
  dimensionality.
\newblock In {\em ICMNS}, 2018.

\bibitem{isola2017image}
Phillip Isola, Jun-Yan Zhu, Tinghui Zhou, and Alexei~A Efros.
\newblock Image-to-image translation with conditional adversarial networks.
\newblock In {\em Proceedings of the IEEE conference on computer vision and
  pattern recognition}, pages 1125--1134, 2017.

\bibitem{kingma2014adam}
Diederik~P Kingma and Jimmy Ba.
\newblock Adam: A method for stochastic optimization.
\newblock {\em arXiv preprint arXiv:1412.6980}, 2014.

\bibitem{kingma2013auto}
Diederik~P Kingma and Max Welling.
\newblock Auto-encoding variational bayes.
\newblock {\em arXiv preprint arXiv:1312.6114}, 2013.

\bibitem{lecun1998gradient}
Yann LeCun, L{\'e}on Bottou, Yoshua Bengio, Patrick Haffner, et~al.
\newblock Gradient-based learning applied to document recognition.
\newblock {\em Proceedings of the IEEE}, 86(11):2278--2324, 1998.

\bibitem{ledig2017photo}
Christian Ledig, Lucas Theis, Ferenc Husz{\'a}r, Jose Caballero, Andrew
  Cunningham, Alejandro Acosta, Andrew Aitken, Alykhan Tejani, Johannes Totz,
  Zehan Wang, et~al.
\newblock Photo-realistic single image super-resolution using a generative
  adversarial network.
\newblock In {\em Proceedings of the IEEE conference on computer vision and
  pattern recognition}, pages 4681--4690, 2017.

\bibitem{lee2008sparse}
Honglak Lee, Chaitanya Ekanadham, and Andrew~Y Ng.
\newblock Sparse deep belief net model for visual area v2.
\newblock In {\em Advances in neural information processing systems}, pages
  873--880, 2008.

\bibitem{lee2009convolutional}
Honglak Lee, Roger Grosse, Rajesh Ranganath, and Andrew~Y Ng.
\newblock Convolutional deep belief networks for scalable unsupervised learning
  of hierarchical representations.
\newblock In {\em Proceedings of the 26th annual international conference on
  machine learning}, pages 609--616. ACM, 2009.

\bibitem{lee2009unsupervised}
Honglak Lee, Peter Pham, Yan Largman, and Andrew~Y Ng.
\newblock Unsupervised feature learning for audio classification using
  convolutional deep belief networks.
\newblock In {\em Advances in neural information processing systems}, pages
  1096--1104, 2009.

\bibitem{makhzani2015adversarial}
Alireza Makhzani, Jonathon Shlens, Navdeep Jaitly, Ian Goodfellow, and Brendan
  Frey.
\newblock Adversarial autoencoders.
\newblock {\em arXiv preprint arXiv:1511.05644}, 2015.

\bibitem{mirza2014conditional}
Mehdi Mirza and Simon Osindero.
\newblock Conditional generative adversarial nets.
\newblock {\em arXiv preprint arXiv:1411.1784}, 2014.

\bibitem{mnih2014neural}
Andriy Mnih and Karol Gregor.
\newblock Neural variational inference and learning in belief networks.
\newblock {\em arXiv preprint arXiv:1402.0030}, 2014.

\bibitem{mohamed2012acoustic}
Abdel-rahman Mohamed, George~E Dahl, and Geoffrey Hinton.
\newblock Acoustic modeling using deep belief networks.
\newblock {\em IEEE Transactions on Audio, Speech, and Language Processing},
  20(1):14--22, 2012.

\bibitem{nair20093d}
Vinod Nair and Geoffrey~E Hinton.
\newblock 3d object recognition with deep belief nets.
\newblock In {\em Advances in neural information processing systems}, pages
  1339--1347, 2009.

\bibitem{nair2010rectified}
Vinod Nair and Geoffrey~E Hinton.
\newblock Rectified linear units improve restricted boltzmann machines.
\newblock In {\em Proceedings of the 27th international conference on machine
  learning (ICML-10)}, pages 807--814, 2010.

\bibitem{neapolitan2004learning}
Richard~E Neapolitan et~al.
\newblock {\em Learning bayesian networks}, volume~38.
\newblock Pearson Prentice Hall Upper Saddle River, NJ, 2004.

\bibitem{nielsen2009bayesian}
Thomas~Dyhre Nielsen and Finn~Verner Jensen.
\newblock {\em Bayesian networks and decision graphs}.
\newblock Springer Science \& Business Media, 2009.

\bibitem{radford2015unsupervised}
Alec Radford, Luke Metz, and Soumith Chintala.
\newblock Unsupervised representation learning with deep convolutional
  generative adversarial networks.
\newblock {\em arXiv preprint arXiv:1511.06434}, 2015.

\bibitem{reed2016generative}
Scott Reed, Zeynep Akata, Xinchen Yan, Lajanugen Logeswaran, Bernt Schiele, and
  Honglak Lee.
\newblock Generative adversarial text to image synthesis.
\newblock {\em arXiv preprint arXiv:1605.05396}, 2016.

\bibitem{salakhutdinov2}
Ruslan Salakhutdinov and Iain Murray.
\newblock On the quantitative analysis of deep belief networks.
\newblock In {\em Proceedings of the 25th international conference on Machine
  learning}, pages 872--879. ACM, 2008.

\bibitem{smolensky1986information}
Paul Smolensky.
\newblock Information processing in dynamical systems: Foundations of harmony
  theory.
\newblock Technical report, Colorado Univ at Boulder Dept of Computer Science,
  1986.

\bibitem{srivastava2012multimodal}
Nitish Srivastava and Ruslan~R Salakhutdinov.
\newblock Multimodal learning with deep boltzmann machines.
\newblock In {\em Advances in neural information processing systems}, pages
  2222--2230, 2012.

\bibitem{stirman2016wide}
Jeffrey~N Stirman, Ikuko~T Smith, Michael~W Kudenov, and Spencer~L Smith.
\newblock Wide field-of-view, multi-region, two-photon imaging of neuronal
  activity in the mammalian brain.
\newblock {\em Nature biotechnology}, 34(8):857, 2016.

\bibitem{sutskever2010convergence}
Ilya Sutskever and Tijmen Tieleman.
\newblock On the convergence properties of contrastive divergence.
\newblock In {\em Proceedings of the thirteenth international conference on
  artificial intelligence and statistics}, pages 789--795, 2010.

\bibitem{teh2001rate}
Yee~Whye Teh and Geoffrey~E Hinton.
\newblock Rate-coded restricted boltzmann machines for face recognition.
\newblock In {\em Advances in neural information processing systems}, pages
  908--914, 2001.

\bibitem{tieleman2008training}
Tijmen Tieleman.
\newblock Training restricted boltzmann machines using approximations to the
  likelihood gradient.
\newblock In {\em Proceedings of the 25th international conference on Machine
  learning}, pages 1064--1071. ACM, 2008.

\bibitem{tieleman2012lecture}
Tijmen Tieleman and Geoffrey Hinton.
\newblock Lecture 6.5-rmsprop: Divide the gradient by a running average of its
  recent magnitude.
\newblock {\em COURSERA: Neural networks for machine learning}, 4(2):26--31,
  2012.

\bibitem{yu2017seqgan}
Lantao Yu, Weinan Zhang, Jun Wang, and Yong Yu.
\newblock Seqgan: Sequence generative adversarial nets with policy gradient.
\newblock In {\em Thirty-First AAAI Conference on Artificial Intelligence},
  2017.

\bibitem{zheng2014eeg}
Wei-Long Zheng, Jia-Yi Zhu, Yong Peng, and Bao-Liang Lu.
\newblock Eeg-based emotion classification using deep belief networks.
\newblock In {\em 2014 IEEE International Conference on Multimedia and Expo
  (ICME)}, pages 1--6. IEEE, 2014.

\end{thebibliography}
\end{document}